\documentclass[10pt,twocolumn,letterpaper]{article}

\usepackage{iccv}
\usepackage{times}
\usepackage{epsfig}
\usepackage{graphicx}
\usepackage{amsmath}
\usepackage{amssymb}

\usepackage{subfigure}
\usepackage{pifont}
\usepackage{color, colortbl}

\usepackage{csquotes}

\usepackage{multirow}
\usepackage{graphicx}
\usepackage{booktabs}

\newcommand*{\twoelementtable}[3][l]%
{%
    \renewcommand{\arraystretch}{0.8}%
    \begin{tabular}[t]{@{}#1@{}}%
        #2\tabularnewline
        #3%
    \end{tabular}%
}
\usepackage{array}
\usepackage{tabularx}

\usepackage{algorithm}
\usepackage{algorithmic}

\usepackage[pagebackref=true,breaklinks=true,letterpaper=true,colorlinks,bookmarks=false]{hyperref}

\iccvfinalcopy 

\def\iccvPaperID{1980} 

\ificcvfinal\fi

\begin{document}

\title{Semi-Supervised Domain Adaptation via Adaptive and Progressive Feature Alignment}

\author{Jiaxing Huang, Dayan Guan, Aoran Xiao, Shijian Lu\thanks{Corresponding author.} \\ 
School of Computer Science Engineering, Nanyang Technological University\\
{\tt\small \{Jiaxing.Huang, Dayan.Guan, Aoran.Xiao, Shijian.Lu\}@ntu.edu.sg}
}

\maketitle
\ificcvfinal\thispagestyle{empty}\fi

\begin{abstract}
Contemporary domain adaptive semantic segmentation aims to address data annotation challenges by assuming that target domains are completely unannotated. However, annotating a few target samples is usually very manageable and worthwhile especially if it improves the adaptation performance substantially. This paper presents SSDAS, a Semi-Supervised Domain Adaptive image Segmentation network that employs a few labeled target samples as anchors for adaptive and progressive feature alignment between labeled source samples and unlabeled target samples. We position the few labeled target samples as references that gauge the similarity between source and target features and guide adaptive inter-domain alignment for learning more similar source features. In addition, we replace the dissimilar source features by high-confidence target features continuously during the iterative training process, which achieves progressive intra-domain alignment between confident and unconfident target features.
Extensive experiments show the proposed SSDAS greatly outperforms a number of baselines, \ie, UDA-based semantic segmentation and SSDA-based image classification.
In addition, SSDAS is complementary and can be easily
incorporated into UDA-based methods with consistent improvements in domain adaptive semantic segmentation.
\end{abstract}

\section{Introduction}
\begin{figure}[t]
\centering
\subfigure{\includegraphics[width=1.0\linewidth]{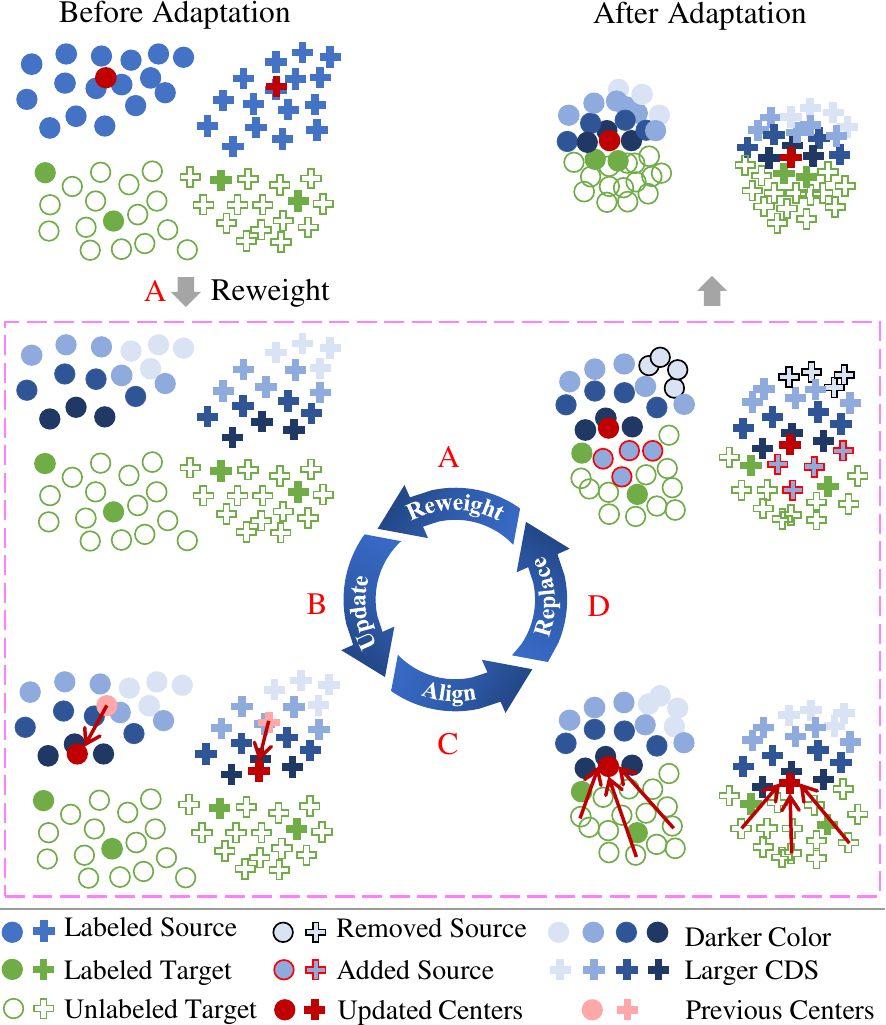}}
\vspace{-2mm}
\caption{
Illustration of our proposed semi-supervised domain adaptive semantic segmentation (SSDAS): With a few labeled target samples as anchors, SSDAS achieves semi-supervised domain adaptation via 4 key processes including \textbf{A}: re-weight source features according to their similarity to the feature of the few labeled target samples, \ie, cross-domain similarity (CDS),
\textbf{B}: update feature centers with newly weighted source features, \textbf{C}: align target features to the updated feature centers, and \textbf{D}: replace dissimilar source features by high-confidence target features. The four processes run iteratively which keep assigning higher weights to target-alike source features and replacing target-unlike source features with confident target features adaptively and progressively (Best viewed in color).
}
\label{fig:intro}
\end{figure}

Domain adaptation has been investigated extensively for the task of semantic segmentation, largely for minimizing the gap between a labeled source domain (typically consisting of synthetic images with automatically generated labels) and an unlabeled target domain and accordingly alleviating the pain point of pixel-level annotation of large amounts of training images. Most existing domain adaptive semantic segmentation works are unsupervised \cite{hoffman2016fcns,kang2019contrastive,kang2018deep,sankaranarayanan2018learning,tsai2018learning,tzeng2017adversarial,luo2019taking,vu2019advent,Chen_2018_CVPR,chen2018domain} which assume a completely unannotated target domain. However, annotating a few target samples is often very manageable and worthwhile especially if it can substantially improve the segmentation performance in the target domain. By managing the amount of labeling in the target domain flexibly, such semi-supervised domain adaptation (SSDA) is also more scalable while facing very diverse domain gaps where unsupervised domain adaptation (UDA) often fails to handle well (e.g. when the domain gap is very large).

SSDA based semantic segmentation has been largely neglected despite its great values in various practical tasks. One intuitive approach to this new problem is to adapt existing UDA methods by including extra supervision from the few labeled target samples in the training process. However, existing UDA methods were designed without considering few-shot labelling in the semi-supervised setup, and recent studies \cite{saito2019semi,kim2020attract} show that this intuitive approach does not perform clearly better (sometimes even worse) than direct training over all labeled source and few-shot target samples in a supervised manner. Considering the similarity between semantic segmentation and image classification, another approach is to adapt the recent SSDA based image classification methods \cite{saito2019semi,kim2020attract} for the semantic segmentation problem. However, the SSDA based image classification methods do not perform well for the dense prediction of image pixels which often experiences much larger and more diverse variations across domains.

We propose SSDAS, a Semi-Supervised Domain Adaptive Segmentation network that positions a few labeled target samples as anchors for effective feature alignment between labeled source samples and unlabeled target samples. We introduce two novel designs for optimal feature alignment across domains. The first design is \textit{adaptive cross-domain alignment} that takes the few labeled target samples as references to gauge the cross-domain similarity (CDS) of each source feature and learn more from target-alike source features. Specifically, we raise (or lower) the weight of source features that are similar (or dissimilar) to the target features adaptively during the domain adaptation process as illustrated in Fig.~\ref{fig:intro}. The second design is \textit{progressive intra-domain alignment} where target-unlike source features are replaced by high-confidence target features progressively and the alignment becomes in-between confident and unconfident target features.
The proposed SSDAS adopts Jigsaw Puzzle classifiers to extract contextual features for domain adaptive semantic segmentation and it can also work with conventional features, more details to be discussed in experiments. 

The contributions of this work can be summarized in three major aspects. \textit{First}, we design SSDAS – an innovative semi-supervised domain adaptive segmentation network that exploits a few labeled target samples for better domain adaptive semantic segmentation. To the best of our knowledge, this is the first work that explores semi-supervised domain adaptation (with few-shot labels) for the classical semantic segmentation task. \textit{Second}, we design two novel feature alignment strategies for semi-supervised domain adaptation problems. The strategies exploit a few labeled target samples and achieve adaptive cross-domain alignment and progressive intra-domain alignment effectively. \textit{Third}, extensive experiments over multiple domain adaptive segmentation tasks show that our proposed SSDAS achieves superior semantic segmentation consistently.

\section{Related Works}

\textbf{Unsupervised Domain Adaptation (UDA)} has been studied extensively for the task of semantic segmentation. Existing UDA-based segmentation methods can be broadly classified into three categories. The first category is \textit{adversarial training} based which utilizes a domain classifier to align source and target distributions in the feature, output or latent space \cite{hoffman2016fcns,long2016unsupervised,tzeng2017adversarial,luo2019taking,tsai2018learning,chen2018road,zhang2017curriculum,saito2017adversarial,saito2018maximum,vu2019advent,tsai2019domain,lee2019sliced,guan2021uncertainty,zhang2021detr,huang2021mlan}. The second category is \textit{image translation} based which translates images from source to target domains to mitigate domain gaps \cite{hoffman2018cycada, sankaranarayanan2018learning,chen2019crdoco, li2019bidirectional,Zhang2019lipreading,hong2018conditional,yang2020fda,huang2021fsdr}. The third category is \textit{self-training} based which utilizes ``pseudo labels" to guide iterative learning over unlabeled target data \cite{zou2018unsupervised,saleh2018effective,zhong2019invariance,zou2019confidence,guan2021scale,huang2021cross}.

\textbf{Semi-supervised Domain Adaptation (SSDA)} assumes the availability of a few labeled target samples beyond labeled source samples and a large amount of unlabeled target samples as in UDA. Several SSDA methods \cite{ao2017fast,donahue2013semi,yao2015semi,saito2019semi,kim2020attract,jiangbidirectional} have been proposed which addresses domain discrepancy by auxiliary constrain optimization~\cite{donahue2013semi}, subspace learning~\cite{yao2015semi}, label smoothing~\cite{ao2017fast}, entropy mini-max~\cite{saito2019semi}, intra-domain discrepancy minimization~\cite{kim2020attract} and bidirectional adversarial training~\cite{jiangbidirectional}. However, most existing SSDA works focus on image classification and the relevant semantic segmentation task involving dense pixel-level predictions is largely neglected. We focus on SSDA-based semantic segmentation (with few-shot target samples) and it is the first effort for this challenging task to the best of our knowledge.

\textbf{Jigsaw Puzzles} is a basic pattern recognition problem that aims to reconstruct an original image from its shuffled patches. In the field of computer science and artificial intelligence, solving jigsaw puzzles \cite{freeman1964apictorial,kosiba1994automatic} has been widely studied for a variety of tasks in image editing \cite{sholomon2014generalized,cho2009patch}, relic re-composition \cite{paumard2018image,brown2008system}, unsupervised visual representation learning \cite{santa2017deeppermnet,doersch2015unsupervised,noroozi2016unsupervised} and generalized network learning \cite{carlucci2019domain}. In this work, we employ Jigsaw Puzzle classifiers to learn contextual visual features (from labeled data) which are very suitable in semantic image segmentation. The learnt features are then exploited to align unlabeled target data for domain adaptive semantic segmentation.

\section{Method}
This section presents our proposed Semi-Supervised Domain Adaptive Segmentation (SSDAS) method. It consists of four subsections that focus on \textit{Task Definition}, \textit{Adaptive Cross-Domain Alignment (ACDA)} that performs adaptive inter-domain alignment according to the similarity of the learned
features, \textit{Progressive Intra-Domain Alignment (PIDA)} that replaces target-unlike source features by high-confidence target features for alignment between confident and unconfident target features, and \textit{Network Training}.

\subsection{Preliminaries}

\textbf{Task Definition:} We focus on the problem of semi-supervised domain adaptation (SSDA) in semantic segmentation. Given the labeled source images \{$X_{s} \subset \mathbb{R}^{H \times W \times 3}$, $Y_{s} \subset (1, C)^{H \times W}$\}, a few labeled target images \{$X_{t} \subset \mathbb{R}^{H \times W \times 3}$, $Y_{t} \subset (1, C)^{H \times W}$\} and a large number of unlabeled target images $X_{t^{u}} \subset \mathbb{R}^{H \times W \times 3}$ ($H$, $W$ and $C$ stands for image height, image width, and the number of semantic classes, respectively), the goal of SSDA-based semantic segmentation is to learn a model $G$ that performs well on unlabeled target-domain data $X_{t^{u}}$. 

Under such data setup, a `S+T' model without any domain adaptation can be derived by training over the labeled source and target images in a fully supervised manner:
\begin{equation}
\begin{split}
\mathcal{L}_{s\text{+}t}(X_{s}, Y_{s}, X_{t}, Y_{t}; G) = l(G(X_{s}), Y_{s}) \\
+ l(G(X_{t}), Y_{t}),    
\end{split}
\label{eq_S+T}
\end{equation}
where $l$ denotes the standard cross entropy loss.

\begin{figure*}[!ht]
\centering
\subfigure{\includegraphics[width=0.99\linewidth]{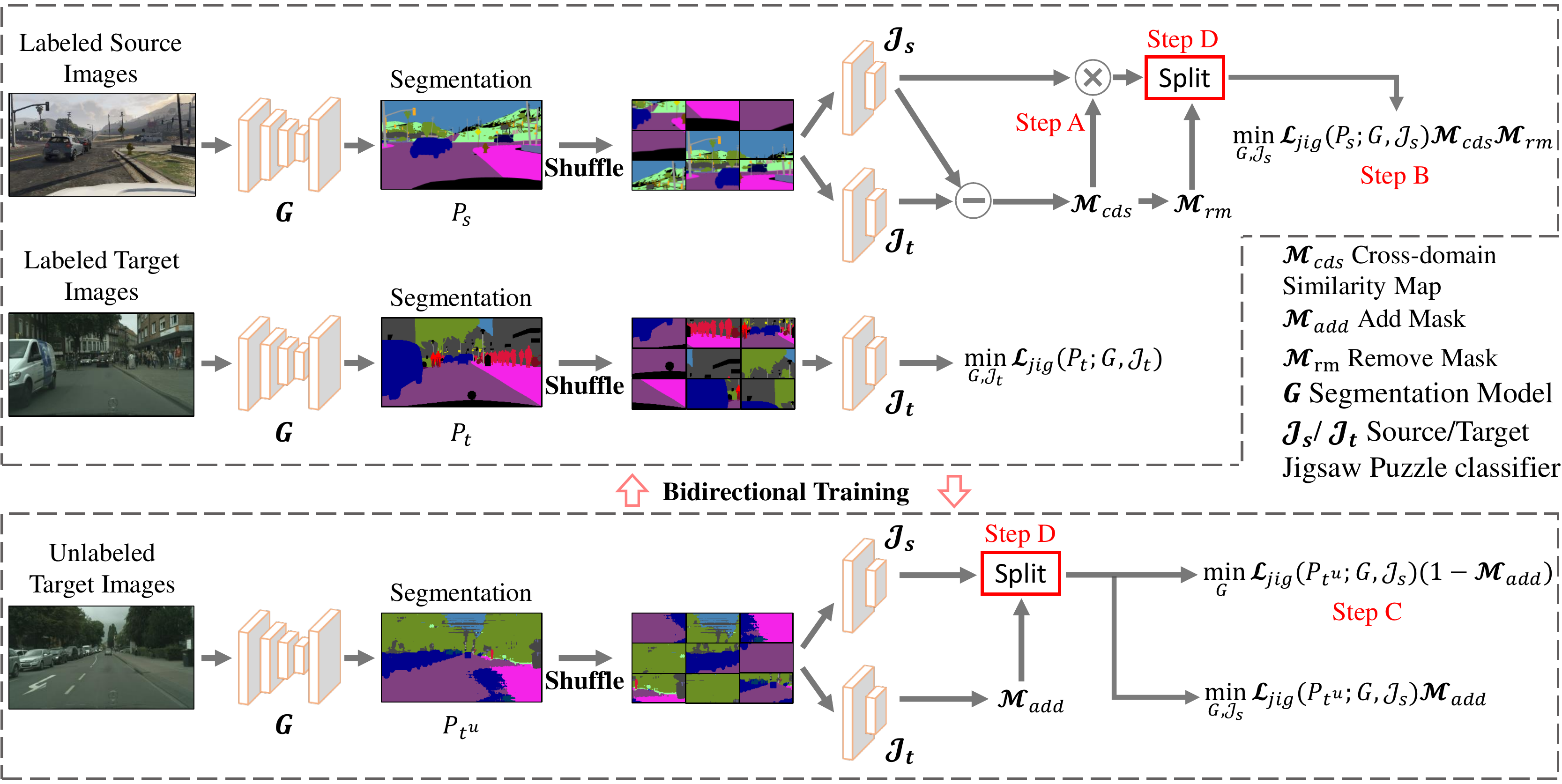}}
\vspace{-2pt}
\caption{
{Overview of the proposed semi-supervised domain adaptive segmentation (SSDAS) network: SSDAS is a bidirectional training framework that consists of two alternative learning processes, namely, labeled data flow (top part) and unlabeled data flow (bottom part). In labeled data flow, we first feed the labeled source and target images ($i.e.$, $X_{s}$ and $X_{t}$) into the segmentation model $G$ to acquire the predicted segmentation maps $P_{s}$ and $P_{t}$, and then employ two Jigsaw Puzzle classifiers ($i.e.$, $\mathcal{J}_{s}$ and $\mathcal{J}_{t}$) to learning context features by solving jigsaw puzzles on $P_{s}$ and $P_{t}$, respectively. In this process, Steps \textcolor{red}{A}, \textcolor{red}{B} and \textcolor{red}{D} are triggered for re-weighting each source feature, re-estimating feature centers with different treatments and removing the dissimilar source features, respectively. In unlabeled data flow ($i.e.$, for $X_{t^{u}}$), we fix the learnt $\mathcal{J}_{s}$ and update $G$ to enforce the the segmentation maps ($i.e.$, $P_{t^{u}}$) to have source-like context features by solving jigsaw puzzles. In this process, Steps \textcolor{red}{C} and \textcolor{red}{D} are triggered for
aligning unconfident target context features to estimated feature centers and adding high-confidence target features to update $G$ and $\mathcal{J}_{s}$, respectively. }
}
\label{fig:stru}
\end{figure*}

\textbf{Context Features Learning.}
Inspired by \cite{noroozi2016unsupervised,carlucci2019domain}, we employ Jigsaw Puzzle classifier $\mathcal{J}$ to learn context features by solving jigsaw puzzles on the predicted segmentation map $P=G(X)$:
\begin{equation}
\mathcal{L}_{jig}(P;G, \mathcal{J}) = l(\mathcal{J}(\text{S}(P, I)), I),
\label{eq_base_jigsaw}
\end{equation}
where `S' stands for a function that decomposes the input image into $n \times n$ patches, and shuffles and re-assigns each patch to one of the $n^{2}$ grid positions (Index $I$ records the original location index of every patch). $\mathcal{J}$ solves jigsaw puzzle ($i.e.$, restoring the shuffled patches) by predicting the original location index ($i.e.$, $I$) of patches.

Note context alignment is a simple yet efficient approach in UDA-based semantic segmentation \cite{tsai2018learning,vu2019advent,luo2019taking,tsai2019domain,huang2020contextual}. For example, \cite{tsai2018learning,vu2019advent,luo2019taking} employ adversarial learning \cite{goodfellow2014generative} to align context features at an image level. \cite{tsai2019domain,huang2020contextual} first sample and cluster patches on labeled data to discover regional context features and then conduct adversarial learning to align the regional context features at a region level. Different from the aforementioned methods, our Jigsaw Puzzle method works at both region and image levels and it learns different types of context features by adjusting the jigsaw puzzle sizes and locations. Please refer to \textcolor{red}{A.5.} in supplementary material for more details.

\subsection{Adaptive Cross-Domain Alignment}
This subsection describes adaptive cross-domain alignment (ACDA) that employs a few labeled target samples as references to re-weight the \textbf{features} of source samples. The features of unlabeled target samples will be aligned to the re-weighted source features for cross-domain alignment.

\textbf{Labeled Data Flow:} For labeled source images \{$X_{s}$, $Y_{s}$\} and labeled target images \{$X_{t}$, $Y_{t}$\}, we first feed them into a segmentation model $G$ to acquire segmentation maps $P_{s} = G(X_{s}) \subset \mathbb{R}^{H \times W \times C}$ and $P_{t} = G(X_{t}) \subset \mathbb{R}^{H \times W \times C}$. We then employ two Jigsaw Puzzle classifiers ($i.e.$, $\mathcal{J}_{s}$ and $\mathcal{J}_{t}$) to learn context features by solving jigsaw puzzles on the predicted segmentation maps:
\begin{equation}
\min_{G, \mathcal{J}_{s}, \mathcal{J}_{t}} \lambda_{j}(\mathcal{L}_{jig}(P_{s};G,\mathcal{J}_{s})\mathcal{M}_{cds} + \mathcal{L}_{jig}(P_{t};G,\mathcal{J}_{t})),
\label{eq_ACDA_label}
\end{equation}
where $\mathcal{M}_{cds} = 1- \mathcal{N}(|\mathcal{J}_{s}(\text{S}(P_{s}, I)) - \mathcal{J}_{t}(\text{S}(P_{s}, I))|)$ is the cross-domain similarity (CDS) map that is computed based on the prediction discrepancy between the source and target Jigsaw Puzzle classifiers. It will be used to re-weight the learning loss of the source context features. $S$ is defined in Eq.~\ref{eq_base_jigsaw} and $\mathcal{N}(x)$ is an unity-based normalization function that brings all values into the range $[0,1]$. As illustrated in Fig.~\ref{fig:intro}, the multiplication with $\mathcal{M}_{cds}$ corresponds to Step \textcolor{red}{A} for source feature re-weighting, and the optimization of this equation corresponds to Step \textcolor{red}{B} for feature center updating.

\textbf{Unlabeled Data Flow:} For unlabeled target images $X_{t^{u}}$, we fix the learnt Jigsaw Puzzle classifier $\mathcal{J}_{s}$ while updating the segmentation model $G$ to solve jigsaw puzzles on the predicted segmentation maps $P_{t^{u}} = G(X_{t^{u}})$. This will enforce the segmentation maps $P_{t^{u}}$ to have source-alike context features:
\begin{equation}
\min_{G} \lambda_{j}(\mathcal{L}_{jig}(P_{t^{u}};G,\mathcal{J}_{s}))
\label{eq_ACDA_unlabel}
\end{equation}
where the optimization of this equation corresponds to Step \textcolor{red}{C} that aligns target features to the updated feature centers as illustrated in Fig.~\ref{fig:intro}.

\subsection{Progressive Intra-Domain Alignment}
This subsection describes our progressive intra-domain alignment (PIDA) that replaces target-unlike source features by high-confidence target features iteratively in training. It leads to intra-domain alignment between confident and unconfident target context features.

\textbf{Labeled Data Flow:} Similar to Eq.~\ref{eq_ACDA_label}, a binary removing mask $\mathcal{M}_{rm}$ is employed to discard target-unlike source features continuously during the iterative training process, where the labeled target data flow remains unchanged:
\begin{equation}
\min_{G, \mathcal{J}_{s}, \mathcal{J}_{t}} \lambda_{j}(\mathcal{L}_{jig}(P_{s};G,\mathcal{J}_{s})\mathcal{M}_{rm} + \mathcal{L}_{jig}(P_{t};G,\mathcal{J}_{t})),
\label{eq_PIDA_label}
\end{equation}
where $\mathcal{M}_{rm}$ is obtained as described in Algorithm~\ref{algorithm_mask_rm}. As illustrated in Fig.~\ref{fig:intro}, the multiplication with the binary mask $\mathcal{M}_{rm}$ corresponds to the Step \textcolor{red}{D} (for removing target-unlike source features), 
and the optimization of this equation correspond to 
the Step \textcolor{red}{B}
.

\textbf{Unlabeled Data Flow:} Based on Eq.~\ref{eq_ACDA_unlabel}, a binary add mask $\mathcal{M}_{add}$ is employed to inject high-confidence target features progressively for updating $G$ and $\mathcal{J}_{s}$, where the unconfident target features will be aligned to confident target features by fixing $\mathcal{J}_{s}$ while updating $G$ during training:
\begin{equation}
\begin{split}
&\min_{G, \mathcal{J}_{s}} \lambda_{j}\mathcal{L}_{jig}(P_{t^{u}};G,\mathcal{J}_{s})\mathcal{M}_{add},\\
&\min_{G} \lambda_{j}\mathcal{L}_{jig}(P_{t^{u}};G,\mathcal{J}_{s})(1-\mathcal{M}_{add}),
\end{split}
\label{eq_PIDA_unlabel}
\end{equation}
where $\mathcal{M}_{add}$ is obtained as described in Algorithm~\ref{algorithm_mask_add}. As illustrated in Fig.~\ref{fig:intro}, the multiplication with the binary mask $\mathcal{M}_{add}$ corresponds to the Step \textcolor{red}{D} (injecting high-confidence target features), and the optimization of the first and second equations refer to Step \textcolor{red}{B} and Step \textcolor{red}{C}, respectively.

\begin{algorithm}[t]
\caption{Determination of $\mathcal{M}_{rm}$ in PIDA.
}\label{algorithm_mask_rm}
  \begin{algorithmic}[1]
\REQUIRE Source images $X_{s}$; $epoch$ denotes the epoch times; segmentation model $G$; jigsaw puzzle classifiers $\mathcal{J}_{s}$ and $\mathcal{J}_{t}$
  \ENSURE $\mathcal{M}_{rm}$
  \STATE Calculate $\mathcal{M}_{cds}$ as in Eq.~\ref{eq_ACDA_label}
  \STATE $\mathcal{M}_{cds}^{sort} = \text{sort}(\mathcal{M}_{cds}, \ \text{order=ascending})$
  \STATE $N =  \text{length}(\mathcal{M}_{cds}^{sort}) \times \frac{epoch}{max\_epoch}$
  \STATE $\text{Thres} =  \mathcal{M}_{cds}^{sort}[N]$
  \STATE $\mathcal{M}_{rm}[\mathcal{M}_{cds}>=\text{Thres}] = 1$
  \STATE $\mathcal{M}_{rm}[\mathcal{M}_{cds}<\text{Thres}] = 0$
  \RETURN $\mathcal{M}_{rm}$
  \end{algorithmic}
\end{algorithm}

\begin{algorithm}[t]
\caption{Determination of $\mathcal{M}_{add}$ in PIDA.
}\label{algorithm_mask_add}
  \begin{algorithmic}[1]
\REQUIRE Unlabeled target images $X_{t^{u}}$; $epoch$ denotes the epoch times; segmentation model $G$; jigsaw puzzle classifier $\mathcal{J}_{t}$
  \ENSURE $\mathcal{M}_{add}$
  \STATE Calculate jigsaw puzzle classification (JPC) probability $P_{t^{u}}^{jig} = \mathcal{J}_{t}(\text{S}(G(X_{t^{u}}), I)$ as in Eq.~\ref{eq_base_jigsaw}
  \STATE Calculate JPC entropy $Ent_{t^{u}}^{jig} = P_{t^{u}}^{jig} \log P_{t^{u}}^{jig}$
  \STATE $Ent_{t^{u}}^{jig\_sort} = \text{sort}(Ent_{t^{u}}^{jig}, \ \text{order=ascending})$
  \STATE $N =  \text{length}(Ent_{t^{u}}^{jig\_sort}) \times \frac{epoch}{max\_epoch}$
  \STATE $\text{Thres} =  Ent_{t^{u}}^{jig\_sort}[N]$
  \STATE $\mathcal{M}_{add}[Ent_{t^{u}}^{jig}<=\text{Thres}] = 1$
  \STATE $\mathcal{M}_{add}[Ent_{t^{u}}^{jig}>\text{Thres}] = 0$
  \RETURN $\mathcal{M}_{add}$
  \end{algorithmic}
\end{algorithm}

\subsection{Network Training}
With ACDA and PIDA as described in Sections 3.3 and 3.4, this subsection presents how ACDA and PIDA work together to achieve our proposed SSDAS. The \textit{labeled data flow} is optimized as follows:
\begin{equation}
\min_{G, \mathcal{J}_{s}, \mathcal{J}_{t}} \lambda_{j}(\mathcal{L}_{jig}(P_{s};G,\mathcal{J}_{s})\mathcal{M}_{cds}\mathcal{M}_{rm} + \mathcal{L}_{jig}(P_{t};G,\mathcal{J}_{t}))
\label{eq_SSDAS_label}
\end{equation}
The optimization function of the \textit{unlabeled data flow} remains unchanged as defined in Eq.~\ref{eq_PIDA_unlabel}

Note all optimization functions in Eqs.~\ref{eq_ACDA_label} - \ref{eq_SSDAS_label} are generic and can be applied for both image-level and region-level alignments. The only difference lies with the input. Specifically, the image-level alignment takes a whole segmentation map as input. But for \textit{region-level alignment}, the segmentation map $P$ ($i.e.$, $P_{s}$, $P_{t}$ and $P_{t^{u}}$) is cropped into $r \times r$ regions which are fed into the extra region-level Jigsaw Puzzle Classifiers ($\mathcal{J}_{s}^{region}$ and $\mathcal{J}_{t}^{region}$) for regional context features learning and alignment. The entire training pipeline is summarized in Algorithm.~\ref{algorithm_SSDAS_entire}.

\begin{algorithm}[t]
\caption{The proposed Semi-Supervised Domain Adaptive Segmentation (SSDAS).
}\label{algorithm_SSDAS_entire}
  \begin{algorithmic}[1]
\REQUIRE Labeled source images $\{X_{s}, Y_{s}\}$ and target images $\{X_{t}, Y_{t}\}$; unlabeled target images $X_{t^{u}}$; $epoch$ denotes the epoch times; segmentation model $G$; jigsaw puzzle classifiers $\mathcal{J}_{s}^{image}$, $\mathcal{J}_{t}^{image}$, $\mathcal{J}_{s}^{region}$ and $\mathcal{J}_{t}^{region}$; region crop size $r$
  \ENSURE Learnt parameters $\theta$ of segmentation model $G$
  \FOR{$epoch = 1$ \textbf{to} $max\_epoch$}
  \STATE \textbf{\emph{Supervised segmentation learning:}}
  \STATE Update $G$ using Eq.~\ref{eq_S+T} with $\{X_{s}, Y_{s}\}$ and $\{X_{t}, Y_{t}\}$
  \STATE \textbf{\emph{Image-level alignment:}} ACDA\&PIDA
  \STATE Calculate segmentation probability map $P_{s} = G(X_{s})$, $P_{t} = G(X_{t})$ and $P_{t^{u}} = G(X_{t^{u}})$
  \STATE Update $G$, $\mathcal{J}_{s}^{image}$ and $\mathcal{J}_{t}^{image}$ using Eq.~\ref{eq_SSDAS_label} and Eq.~\ref{eq_PIDA_unlabel} with $P_{s}$, $P_{t}$ and $P_{t^{u}}$
  \STATE \textbf{\emph{Region-level alignment:}} ACDA\&PIDA
  \STATE Crop maps into $r \times r$ regions $P_{s}^{region} = \{P_{s}^{0},P_{s}^{1},...,P_{s}^{r^{2}}\}$, $P_{t}^{region} = \{P_{t}^{0},P_{t}^{1},...,P_{t}^{r^{2}}\}$ and $P_{t^{u}}^{region} = \{P_{t^{u}}^{0},P_{t^{u}}^{1},...,P_{t^{u}}^{r^{2}}\}$
  \STATE Update $G$, $\mathcal{J}_{s}^{region}$ and $\mathcal{J}_{t}^{region}$ using Eq.~\ref{eq_SSDAS_label} and Eq.~\ref{eq_PIDA_unlabel} with $P_{s}^{region}$, $P_{t}^{region}$ and $P_{t^{u}}^{region}$
  \ENDFOR
  \RETURN $G$
  \end{algorithmic}
\end{algorithm}

\renewcommand\arraystretch{1.1}
\begin{table}[t]
\centering
\resizebox{\columnwidth}{!}{
\begin{tabular}{cccc|c}
\hline
\multicolumn{2}{c|}{\multirow{1}{*}{Image-level Alignment}} &\multicolumn{2}{c|}{\multirow{1}{*}{Region-level Alignment}} & \multicolumn{1}{c}{\multirow{2}{*}{mIoU}}
\\\cline{1-4}
ACDA &\multicolumn{1}{c|}{PIDA} &ACDA &\multicolumn{1}{c|}{PIDA}
\\\hline
& & & &37.9\\\hline
\multicolumn{1}{c}{\checkmark} & & & &43.6\\
&\multicolumn{1}{c}{\checkmark}  & & &41.7\\
\multicolumn{1}{c}{\checkmark} &\multicolumn{1}{c}{\checkmark}  & & &45.1\\\hline
& &\multicolumn{1}{c}{\checkmark} & &44.8\\
& & &\multicolumn{1}{c|}{\checkmark} &41.9\\
& &\multicolumn{1}{c}{\checkmark} &\multicolumn{1}{c|}{\checkmark} &46.2\\\hline
\multicolumn{1}{c}{\checkmark}& &\multicolumn{1}{c}{\checkmark} & &46.3\\
&\multicolumn{1}{c}{\checkmark} & &\multicolumn{1}{c|}{\checkmark} &44.5\\
\multicolumn{1}{c}{\checkmark} &\multicolumn{1}{c}{\checkmark} &\multicolumn{1}{c}{\checkmark} &\multicolumn{1}{c|}{\checkmark} &\textbf{48.5}\\
\hline
\end{tabular}}
\caption{
Ablation study of SSDAS over semi-supervised domain adaptive segmentation task GTA $\rightarrow$ Cityscapes. The \textit{1st} row shows ``S+T" model that is trained with supervised segmentation loss with labeled source and target samples only (without any alignment) as defined in Eq.~\ref{eq_S+T}.
The backbone is ResNet-101 and the setting is one-shot as evaluated in mIoU.
}
\label{tab:abla}
\end{table}

\begin{table*}[t]
	\centering
	\resizebox{\linewidth}{!}{
	\begin{tabular}{c|c|ccccccccccccccccccc|c}
		\hline
		Setting &Method  & Road & SW & Build & Wall & Fence & Pole & TL & TS & Veg. & Terrain & Sky & PR & Rider & Car & Truck & Bus & Train & Motor & Bike & mIoU\\
		\hline
		\multirow{10}{0.1\linewidth}{\centering{1-shot}} &S+T &{82.1}	&22.8	&72.8	&20.3	&22.0	&26.7	&31.1	&11.8	&75.7	&24.0	&76.8	&56.2	&23.2	&71.4	&21.6	&20.7	&0.1	&28.5	&32.5	&37.9\\
		&AdaptSeg~\cite{tsai2018learning} &{87.0}	&32.8	&80.5	&22.1	&22.6	&27.7	&34.3	&25.4	&83.4	&31.5	&81.7	&59.5	&18.6	&76.0	&35.4	&42.9	&1.6	&28.0	&27.1	&43.1\\
		&ADVENT~\cite{vu2019advent} &{92.3}	&52.8	&81.9	&30.0	&25.6	&28.8	&35.4	&24.0	&84.9	&39.1	&79.4	&56.1	&21.4	&86.3	&31.1	&33.5	&1.2	&30.8	&11.1	&44.5\\
		&CRST~\cite{zou2019confidence} &{91.5}	&51.7	&81.6	&28.3	&\textbf{28.5}	&\textbf{43.9}	&45.0	&26.7	&85.2	&35.4	&65.0	&\textbf{68.8}	&29.2	&85.5	&32.8	&28.4	&1.3	&31.0	&\textbf{42.0}	&47.5\\
		&FDA~\cite{yang2020fda} &90.5	&40.9	&80.2	&26.0	&25.0	&31.1	&32.6	&34.5	&80.3	&33.7	&79.5	&54.0	&30.2	&83.8	&35.0	&41.2	&11.8	&21.1	&30.7	&45.4\\
		&IDA~\cite{pan2020unsupervised} &{92.7}	&53.2	&82.9	&31.9	&19.7	&29.9	&35.6	&22.5	&86.2	&45.8	&\textbf{82.4}	&57.8	&26.6	&87.9	&33.2	&42.7	&1.1	&\textbf{32.9}	&24.3	&46.8\\\cline{2-22}
		&\textbf{SSDAS} &92.9	&53.9	&82.1	&31.5	&24.0	&36.4	&40.6	&33.8	&84.5	&44.9	&69.6	&60.5	&25.4	&85.1	&\textbf{46.9}	&52.4	&2.7	&19.2	&34.3	&48.5\\
		&\textbf{+ADVENT} &{93.2}	&54.1	&82.4	&32.8	&25.4	&34.0	&39.5	&36.7	&85.0	&45.7	&80.4	&60.7	&26.8	&84.3	&42.5	&\textbf{52.5}	&3.1	&28.5	&29.4	&49.3\\
		&\textbf{+CRST} &\textbf{{94.1}}	&60.8	&\textbf{83.9}	&\textbf{35.3}	&17.5	&40.1	&\textbf{50.6}	&\textbf{47.1}	&85.2	&45.6	&81.2	&66.3	&26.9	&\textbf{88.6}	&33.7	&49.1	&10.7	&22.7	&35.8	&\textbf{51.3}\\
		&\textbf{+FDA} &93.9	&52.0	&83.3	&28.2	&26.8	&38.9	&37.6	&38.6	&82.7	&40.7	&81.4	&54.4	&\textbf{32.9}	&86.6	&42.1	&49.3	&\textbf{14.8}	&22.8	&34.9	&49.6\\
		&\textbf{+IDA} &93.7	&\textbf{56.7}	&83.8	&33.9	&25.7	&35.8	&40.3	&38.2	&\textbf{86.6}	&\textbf{46.1}	&81.2	&60.0	&25.1	&87.8	&34.1	&47.1	&3.3	&26.1	&34.8	&49.5\\
		\hline
		\multirow{10}{0.1\linewidth}{\centering{3-shot}} &S+T &{73.2}	&29.9	&75.4	&17.6	&20.4	&30.5	&34.7	&24.5	&80.8	&26.0	&76.0	&58.1	&28.1	&45.1	&34.8	&34.1	&0.6	&26.9	&37.0	&39.7\\
		&AdaptSeg~\cite{tsai2018learning} &86.6	&43.3	&80.7	&22.1	&21.9	&26.1	&33.4	&27.0	&82.8	&28.8	&80.6	&58.1	&26.1	&77.9	&37.2	&42.0	&1.1	&24.8	&29.6	&43.7\\
		&ADVENT~\cite{vu2019advent} &92.3	&51.1	&81.8	&29.7	&23.7	&31.9	&32.7	&18.2	&84.4	&35.5	&75.6	&57.8	&21.5	&86.7	&35.3	&48.0	&0.7	&28.4	&19.7	&45.0\\
		&CRST~\cite{zou2019confidence} &92.2	&52.1	&81.6	&24.7	&27.5	&41.0	&\textbf{45.8}	&27.6	&83.6	&34.2	&76.4	&63.6	&22.4	&86.3	&33.1	&48.0	&5.9	&28.3	&38.2	&48.0\\
		&FDA~\cite{yang2020fda} &91.6	&45.5	&82.4	&25.8	&25.5	&30.7	&34.2	&30.2	&82.7	&29.2	&79.9	&60.3	&28.1	&87.0	&33.2	&37.4	&8.8	&22.6	&34.8	&45.8\\
		&IDA~\cite{pan2020unsupervised} &92.6	&52.2	&83.3	&30.3	&26.7	&33.0	&35.7	&25.2	&85.2	&42.9	&79.5	&59.2	&27.0	&87.3	&37.0	&48.9	&4.5	&30.2	&12.5 &47.0\\\cline{2-22}
		&\textbf{SSDAS} &92.4	&52.6	&83.7	&27.0	&20.9	&37.5	&41.5	&36.3	&85.3	&42.2	&78.7	&63.1	&32.0	&86.8	&45.7	&49.6	&5.9	&20.2	&42.3	&49.7\\
		&\textbf{+ADVENT} &\textbf{93.9}	&52.0	&83.3	&29.2	&26.8	&38.9	&40.6	&38.6	&\textbf{85.7}	&41.7	&81.4	&59.4	&32.9	&86.6	&42.1	&49.3	&\textbf{14.8}	&22.8	&34.9	&50.3\\
		&\textbf{+CRST} &93.6	&\textbf{58.1}	&83.5	&\textbf{32.1}	&\textbf{28.1}	&\textbf{42.7}	&43.8	&\textbf{41.4}	&85.5	&42.1	&81.8	&\textbf{64.9}	&\textbf{33.7}	&\textbf{87.9}	&\textbf{46.8}	&49.5	&13.5	&\textbf{31.9}	&43.3	&\textbf{52.9}\\
		&\textbf{+FDA} &93.3	&49.8	&83.1	&31.3	&25.7	&37.2	&39.1	&38.0	&85.6	&\textbf{43.2}	&81.4	&61.9	&30.1	&86.4	&43.9	&\textbf{50.3}	&9.4	&25.6	&\textbf{43.9}	&50.5\\
		&\textbf{+IDA} &93.6	&56.3	&\textbf{84.3}	&25.3	&21.6	&38.2	&41.6	&38.7	&85.6	&40.0	&\textbf{82.5}	&61.6	&29.4	&86.0	&42.2	&48.6	&7.6	&29.9	&42.6	&50.3\\
		\hline
	\end{tabular}
	}
	\caption{
	Comparing SSDAS with state-of-the-art semantic segmentation methods: For semi-supervised domain adaptive semantic segmentation task GTA $\rightarrow$ \textbf{City}scapes, the proposed SSDAS consistently outperforms all state-of-the-art UDA methods that are adapted for the SSDA task. In addition, SSDAS is clearly complementary to all adapted UDA methods with clear performance gains. The backbone is ResNet-101 for all compared methods, and the setting includes 1-shot and 3-shot.
	}
	\label{table:gta2city}
\end{table*}

\begin{table*}[t]
	\centering
	\resizebox{\linewidth}{!}{
	\begin{tabular}{c|c|cccccccccccccccc|c|c}
		\hline
		Setting &Method  & Road & SW & Build & Wall\textsuperscript{*} & Fence\textsuperscript{*} & Pole\textsuperscript{*} & TL & TS & Veg. & Sky & PR & Rider & Car & Bus & Motor & Bike & mIoU & mIoU\textsuperscript{*}\\
		\hline
		\multirow{10}{0.1\linewidth}{\centering{1-shot}} &S+T &56.3	&17.6	&76.3	&9.9	&2.1	&28.3	&14.3	&13.8	&80.0	&80.9	&51.2	&14.2	&43.1	&22.1	&19.0	&24.6	&34.6	&39.5\\
		&AdaptSeg~\cite{tsai2018learning} &84.2	&41.4	&78.0	&10.1	&0.5	&27.8	&6.1	&11.7	&81.8	&79.5	&54.3	&21.9	&70.8	&35.3	&12.3	&32.8	&40.5	&46.9\\
		&ADVENT~\cite{vu2019advent} &87.2	&45.3	&78.7	&9.2	&2.8	&23.6	&6.7	&14.7	&80.9	&83.3	&58.8	&22.0	&71.3	&31.4	&10.8	&35.1	&41.4	&48.2\\
		&CRST~\cite{zou2019confidence} &69.9	&31.8	&74.9	&14.8	&3.6	&37.0	&22.5	&29.5	&81.6	&79.1	&58.2	&28.8	&83.6	&27.2	&22.9	&46.6	&44.5	&50.5\\
		&FDA~\cite{yang2020fda} &78.2	&32.5	&73.0	&11.2	&2.4	&27.8	&16.1	&17.5	&80.0	&82.0	&52.5	&24.7	&74.3	&34.3	&20.0	&39.8	&41.6	&48.1\\
		&IDA~\cite{pan2020unsupervised} &84.7	&38.3	&78.8	&10.5	&2.9	&27.0	&13.6	&12.3	&80.5	&\textbf{83.8}	&57.7	&22.9	&72.0	&38.0	&20.7	&35.8	&42.5	&49.2 \\\cline{2-20}
		&\textbf{SSDAS} &87.6	&44.4	&79.1	&13.2	&3.2	&29.8	&14.1	&18.4	&81.0	&80.8	&58.0	&26.3	&77.3	&39.7	&19.4	&38.6	&44.4	&51.1\\
		&\textbf{+ADVENT} &\textbf{88.6}	&\textbf{46.2}	&78.8	&14.1	&2.2	&28.9	&16.8	&21.5	&80.9	&82.5	&58.9	&25.5	&78.7	&37.5	&20.3	&41.9	&45.2	&52.2\\
		&\textbf{+CRST} &86.8	&45.7	&79.4	&\textbf{15.2}	&3.1	&\textbf{39.7}	&\textbf{24.0}	&\textbf{31.9}	&\textbf{82.5}	&79.6	&57.7	&\textbf{29.1}	&\textbf{84.4}	&\textbf{41.5}	&\textbf{25.2}	&\textbf{48.7}	&\textbf{48.4}	&\textbf{55.1}\\
		&\textbf{+FDA} &86.3	&43.6	&78.1	&14.5	&\textbf{4.1}	&32.3	&18.6	&20.0	&80.1	&82.8	&57.4	&27.9	&79.8	&40.1	&22.8	&43.9	&45.8	&52.4\\
		&\textbf{+IDA} &87.3	&44.9	&\textbf{80.2}	&12.2	&2.1	&30.8	&12.3	&21.3	&81.8	&83.7	&\textbf{59.4}	&27.1	&75.4	&40.9	&21.8	&41.9	&45.2	&52.2\\
		\hline
		\multirow{10}{0.1\linewidth}{\centering{3-shot}} &S+T &58.4	&24.3	&77.3	&9.8	&2.4	&27.4	&12.6	&16.1	&77.7	&78.7	&51.6	&18.7	&40.0	&28.7	&17.0	&30.3	&35.7	&40.9\\
		&AdaptSeg~\cite{tsai2018learning} &84.7	&38.9	&78.1	&12.6	&2.1	&28.2	&8.8	&12.3	&80.9	&80.1	&55.4	&19.5	&72.7	&35.6	&14.6	&32.2	&41.0	&47.2\\
		&ADVENT~\cite{vu2019advent} &86.8	&44.3	&79.2	&14.1	&4.1	&28.6	&13.2	&18.8	&81.4	&82.7	&56.7	&21.0	&77.2	&33.1	&12.6	&27.1	&42.6	&48.8\\
		&CRST~\cite{zou2019confidence} &72.6	&36.3	&76.9	&15.2	&4.1	&37.5	&21.1	&28.6	&81.9	&82.5	&57.8	&27.4	&82.4	&30.5	&20.1	&42.2	&44.8	&50.8\\
		&FDA~\cite{yang2020fda} &81.6	&36.6	&74.2	&15.6	&3.1	&26.5	&18.1	&19.9	&82.3	&83.0	&55.2	&18.6	&80.7	&27.5	&19.9	&34.5	&42.3	&48.6\\
		&IDA~\cite{pan2020unsupervised} &85.5	&40.0	&78.6	&14.2	&3.1	&25.3	&16.3	&18.3	&80.5	&82.2	&54.9	&19.4	&75.9	&39.6	&19.8	&32.5	&42.9	&49.5
		\\\cline{2-20}
		&\textbf{SSDAS} &88.5	&45.1	&78.2	&15.0	&3.0	&29.2	&19.9	&21.3	&80.8	&82.6	&58.5	&26.0	&76.0	&37.9	&21.4	&40.0	&45.2	&52.0\\
		&\textbf{+ADVENT} &88.7	&46.6	&79.9	&16.9	&4.1	&32.6	&20.2	&20.3	&81.9	&83.5	&57.5	&26.4	&80.7	&38.1	&22.3	&38.7	&46.2	&52.7\\
		&\textbf{+CRST} &\textbf{89.1}	&\textbf{46.7}	&\textbf{80.2}	&16.6	&4.2	&\textbf{40.5}	&\textbf{23.8}	&\textbf{32.9}	&82.8	&83.3	&\textbf{59.7}	&\textbf{29.3}	&\textbf{84.6}	&\textbf{41.0}	&\textbf{24.1}	&\textbf{47.3}	&\textbf{49.1}	&\textbf{55.8}\\
		&\textbf{+FDA} &88.0	&43.7	&78.0	&\textbf{17.6}	&\textbf{5.2}	&31.8	&22.9	&20.4	&\textbf{83.4}	&\textbf{83.8}	&56.2	&27.7	&81.0	&38.1	&22.6	&42.9	&46.5	&53.0\\
		&\textbf{+IDA} &87.9	&46.3	&79.4	&16.4	&4.0	&30.7	&22.0	&20.1	&82.6	&83.1	&57.3	&27.9	&80.8	&40.2	&20.4	&38.9	&46.1	&52.8\\
		\hline
	\end{tabular}
	}
\caption{Comparing SSDAS with state-of-the-art semantic segmentation methods: For semi-supervised domain adaptive semantic segmentation task SYNTHIA $\rightarrow$ Cityscapes, the proposed SSDAS consistently outperforms all state-of-the-art UDA methods that are adapted for the SSDA task. In addition, SSDAS is clearly complementary to all adapted UDA methods with clear performance gains. The backbone is ResNet-101 for all compared methods, and the setting includes 1-shot and 3-shot.}
\label{table:synthia2city}
\end{table*}

\begin{table}[h]
\centering
\vspace{1mm}
\footnotesize
\begin{tabular}{l|c|cc} \hline
 \multirow{2}{*}{Network} & \multirow{2}{*}{Method}       &\multicolumn{2}{c}{GTA $\rightarrow$ City}\\
 &        & 1-shot & 3-shot\\\hline
\multirow{4}{*}{ResNet-101} &S+T  &37.9& 39.7\\
&ENT~\cite{grandvalet2005semi} &42.8 &43.5\\
&MME~\cite{saito2019semi} &43.2	&43.8\\
&\textbf{SSDAS} &\textbf{48.5} &\textbf{49.7}\\
\hline
\end{tabular}
\caption{
Comparing our SSDAS with state-of-the-art SSDA classification methods (in mIoU): For domain adaptive semantic segmentation task GTA $\rightarrow$ \textbf{City}scapes, SSDAS outperforms the state-of-the-art by large margins consistently for both 1-shot and 3-shot.}
\label{tab:SSDA_classification_comp1}
\end{table}

\begin{table*}[t]
\begin{center}
\scalebox{0.82}{
\begin{tabular}{c|l|cccccccccccc|c}
\hline
Network& Method       &R to C& R to P & R to A & P to R & P to C & P to A & A to P & A to C & A to R & C to R & C to A & C to P & MEAN \\\hline

\multirow{5}{*}{ResNet-34} & S+T &  55.7  & 80.8 & 67.8 & 73.1 & 53.8 & 63.5 & 73.1 & 54.0 & 74.2 & 68.3 & 57.6 & 72.3 & 66.2\\
&ENT~\cite{grandvalet2005semi} &  62.6 & 85.7 & 70.2 & 79.9 & 60.5 & 63.9 & 79.5 & 61.3 & 79.1 & 76.4 & 64.7 & 79.1 & 71.9\\
& MME~\cite{saito2019semi} &  64.6 & 85.5 & 71.3 & 80.1 & 64.6 & 65.5 & 79.0 & 63.6 & 79.7 & 76.6 & \textbf{67.2} & 79.3 & 73.1 \\
& APE~\cite{kim2020attract} &  66.4 &  86.2 &  73.4 &  82.0 &  65.2 &  66.1 &  81.1 &  \textbf{63.9} &  80.2 &  76.8 & 66.6 &  79.9 & 74.0\\
& \textbf{SSDAS} &\textbf{69.1}	&\textbf{86.9}	&\textbf{76.2}	&\textbf{83.4}	&\textbf{66.8}	&\textbf{67.5}	&\textbf{83.5}	&63.8	&\textbf{82.3}	&\textbf{77.9}	&67.0	&\textbf{81.1}	&\textbf{75.5}\\
\hline
\end{tabular}}
\end{center}
\caption{
Comparing SSDAS with state-of-the-art SSDA classification methods: For domain adaptive image classification task (3-shot), the proposed SSDAS outperforms the state-of-the-art clearly in most of 12 adaptation scenarios in Office-home dataset.}
\label{tab:SSDA_classification_comp2}
\end{table*}

\begin{figure*}[t]
\begin{tabular}{p{3cm}<{\centering} p{3cm}<{\centering} p{3cm}<{\centering} p{3cm}<{\centering} p{3cm}<{\centering}}
\raisebox{-0.5\height}{Target Image}
 & \raisebox{-0.5\height}{S+T}
& \raisebox{-0.5\height}{IDA~\cite{pan2020unsupervised}}
& \raisebox{-0.5\height}{\textbf{SSDAS (Ours)}}
& \raisebox{-0.5\height}{Ground Truth}
\\
\raisebox{-0.5\height}{\includegraphics[width=1.1\linewidth,height=0.55\linewidth]{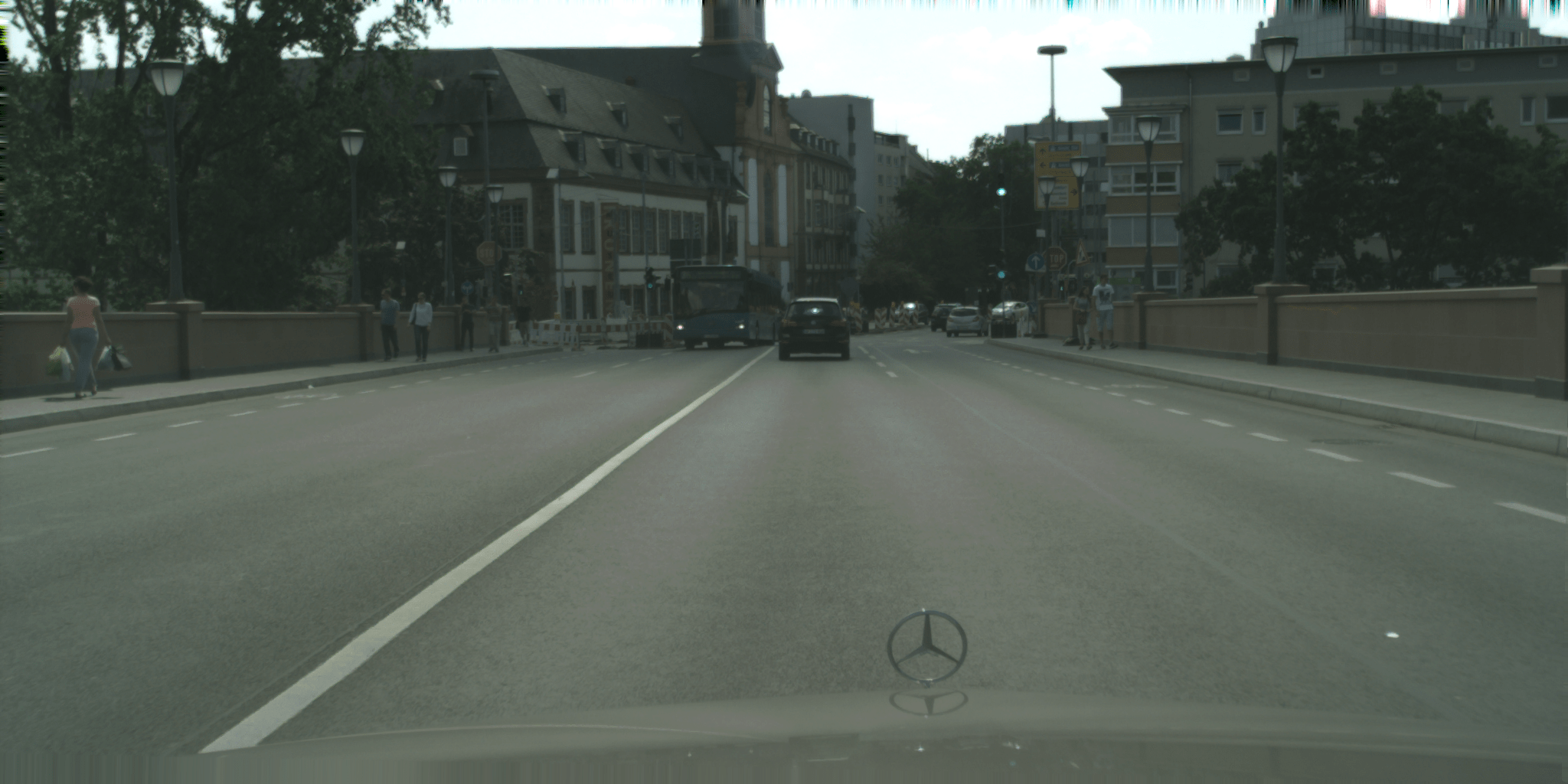}}
 & \raisebox{-0.5\height}{\includegraphics[width=1.1\linewidth,height=0.55\linewidth]{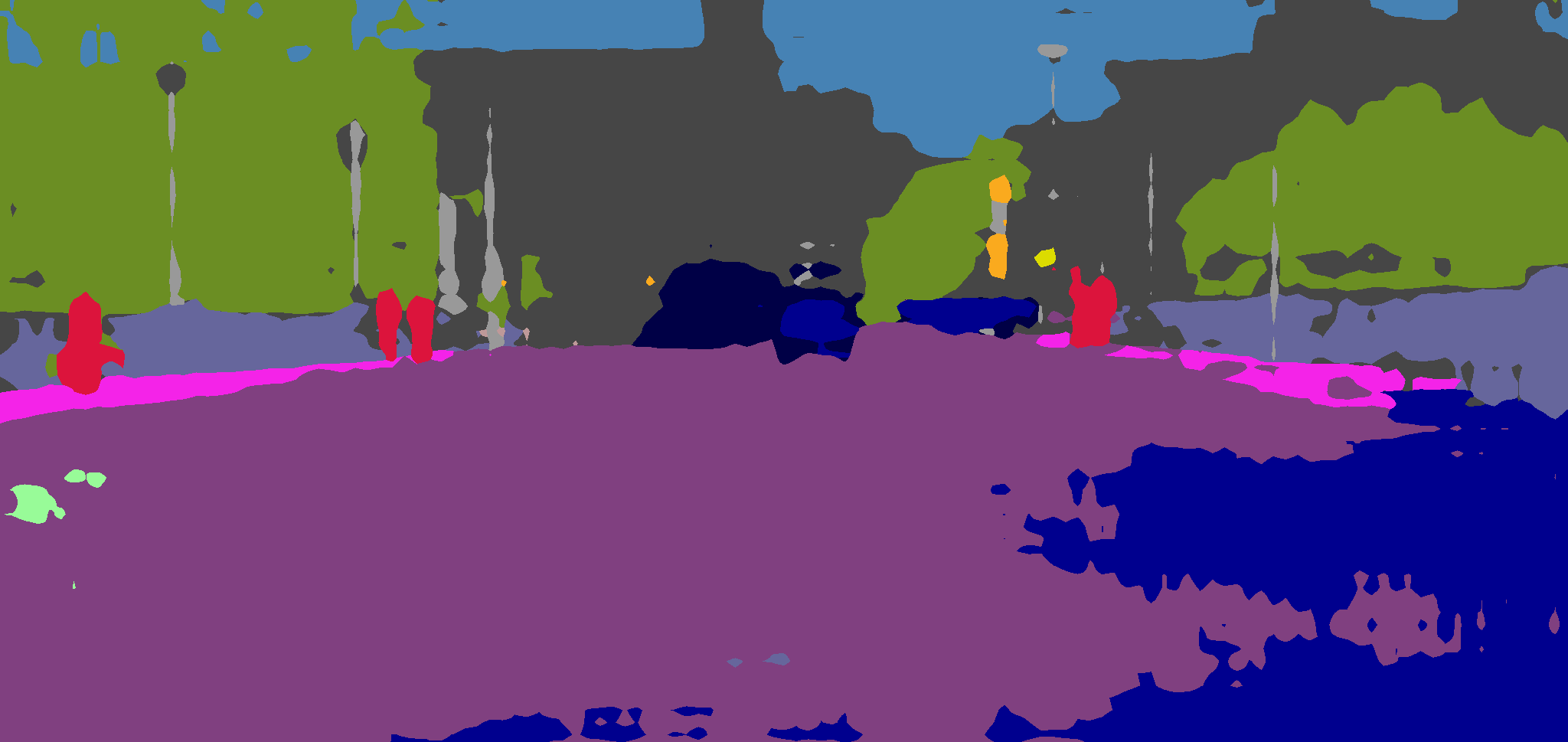}}
& \raisebox{-0.5\height}{\includegraphics[width=1.1\linewidth,height=0.55\linewidth]{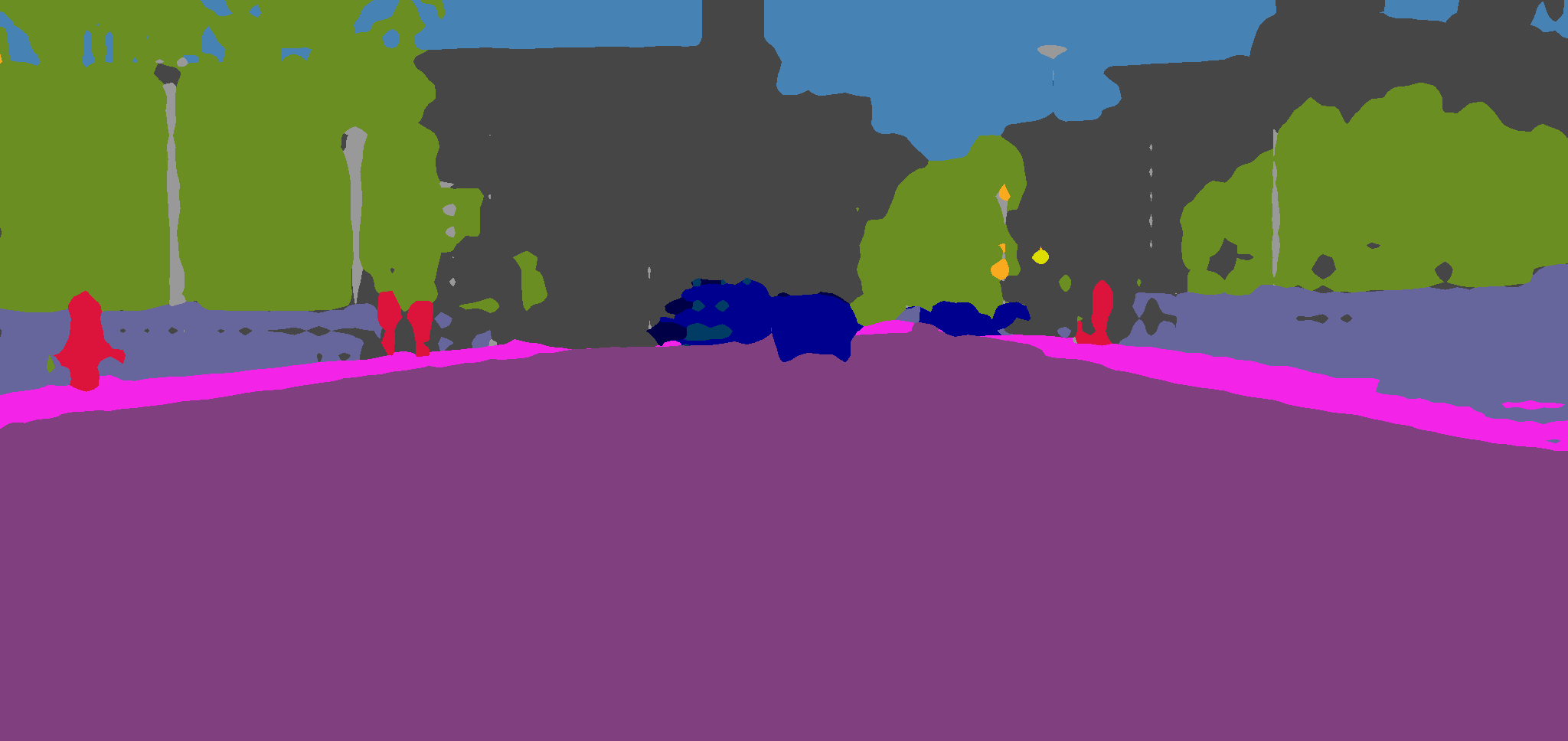}}
& \raisebox{-0.5\height}{\includegraphics[width=1.1\linewidth,height=0.55\linewidth]{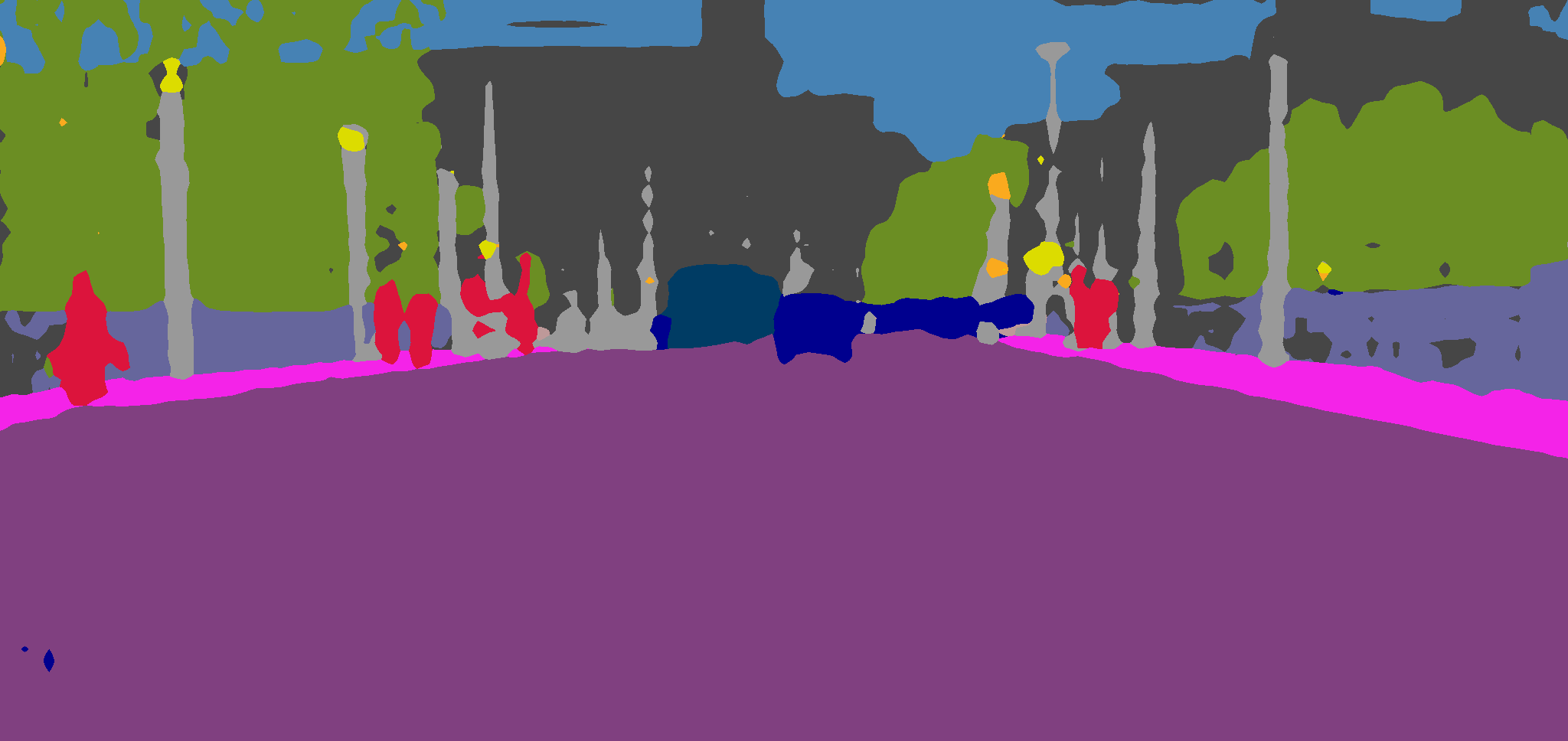}}
& \raisebox{-0.5\height}{\includegraphics[width=1.1\linewidth,height=0.55\linewidth]{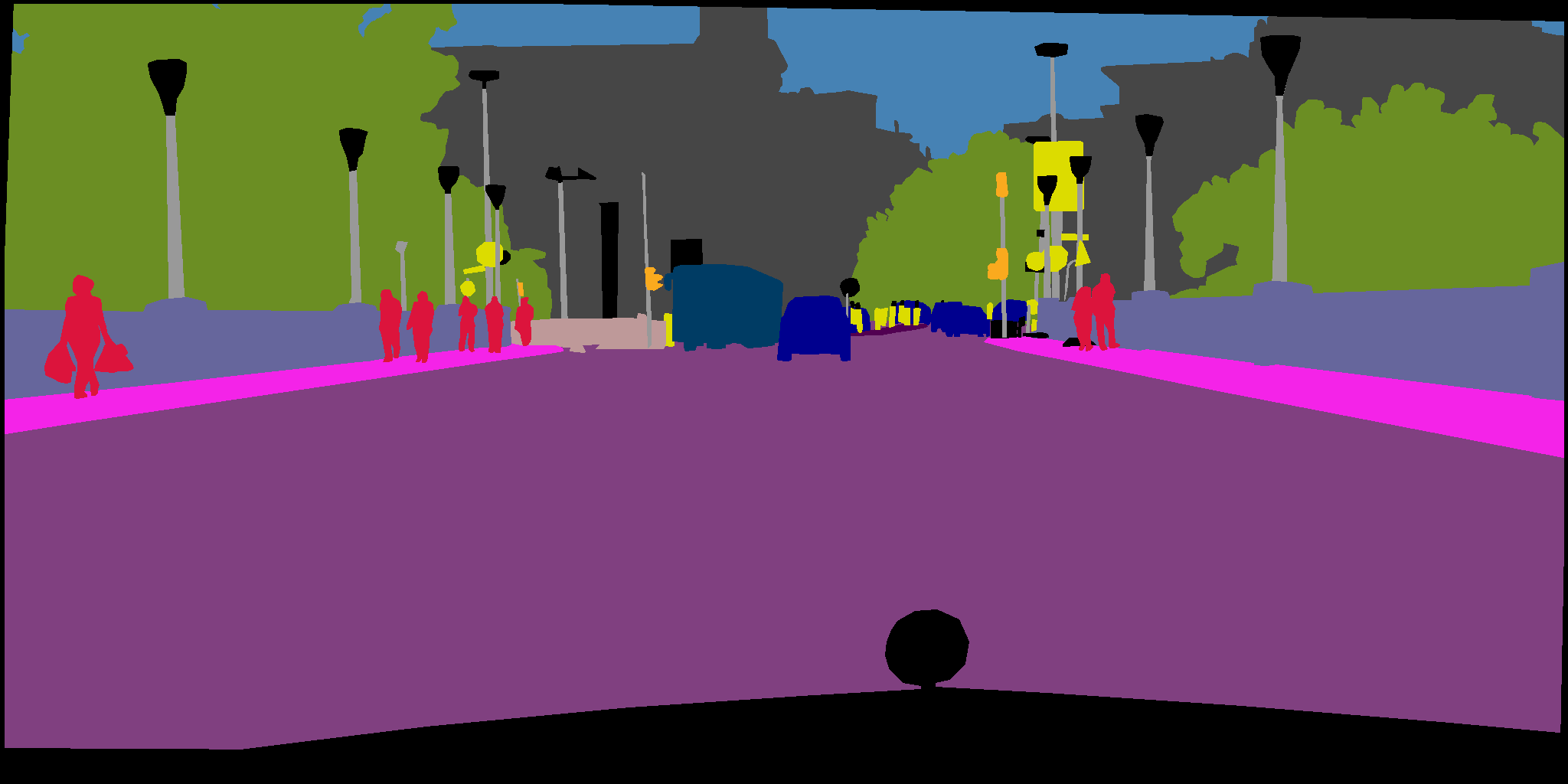}}
\vspace{-2.5 pt}
\\
\end{tabular}
\begin{tabular}{p{5.18cm}<{\centering} p{5.18cm}<{\centering} p{5.18cm}<{\centering}}
\raisebox{-0.5\height}{\footnotesize{$\sigma_{w}^{2}=530.11$, $\sigma_{b}^{2}=42.60$}}
 & \raisebox{-0.5\height}{\footnotesize{$\sigma_{w}^{2}=306.28$, $\sigma_{b}^{2}=46.25$}}
& \raisebox{-0.5\height}{\footnotesize{$\sigma_{w}^{2}=196.75$, $\sigma_{b}^{2}=57.71$}}
\vspace{-10 pt}
\\
\raisebox{-0.5\height}{\includegraphics[width=1\linewidth,height=0.8\linewidth]{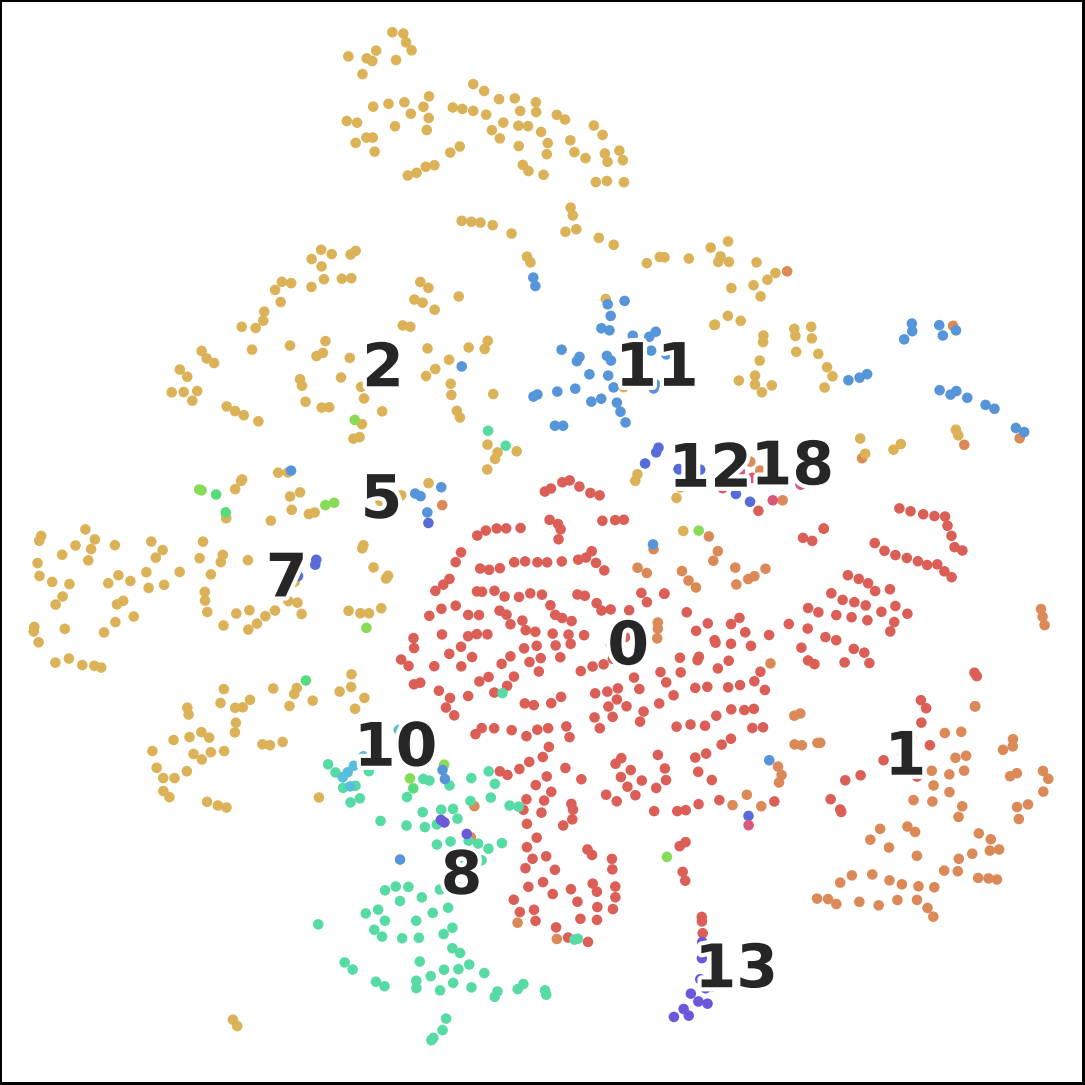}}
 & \raisebox{-0.5\height}{\includegraphics[width=1\linewidth,height=0.8\linewidth]{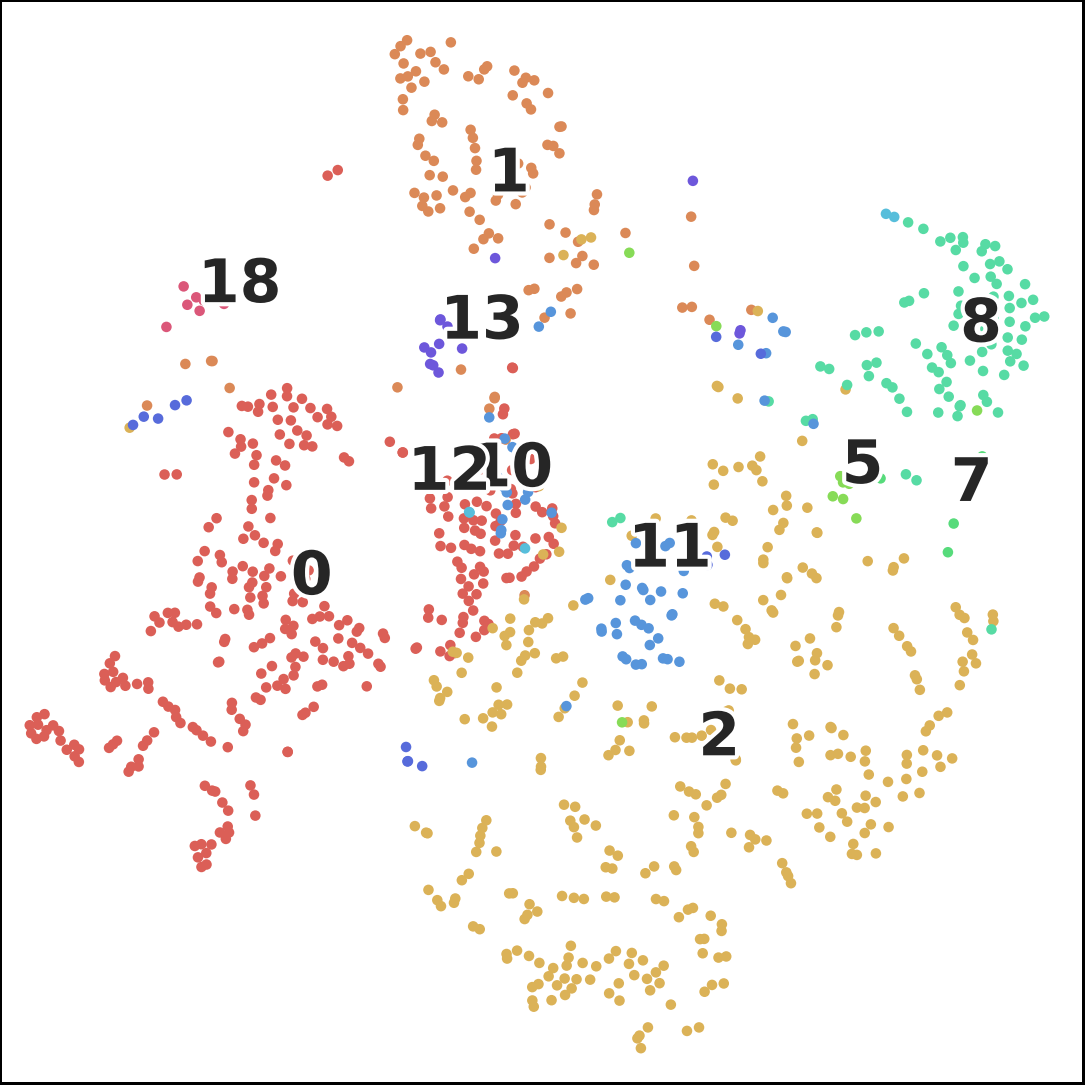}}
& \raisebox{-0.5\height}{\includegraphics[width=1\linewidth,height=0.8\linewidth]{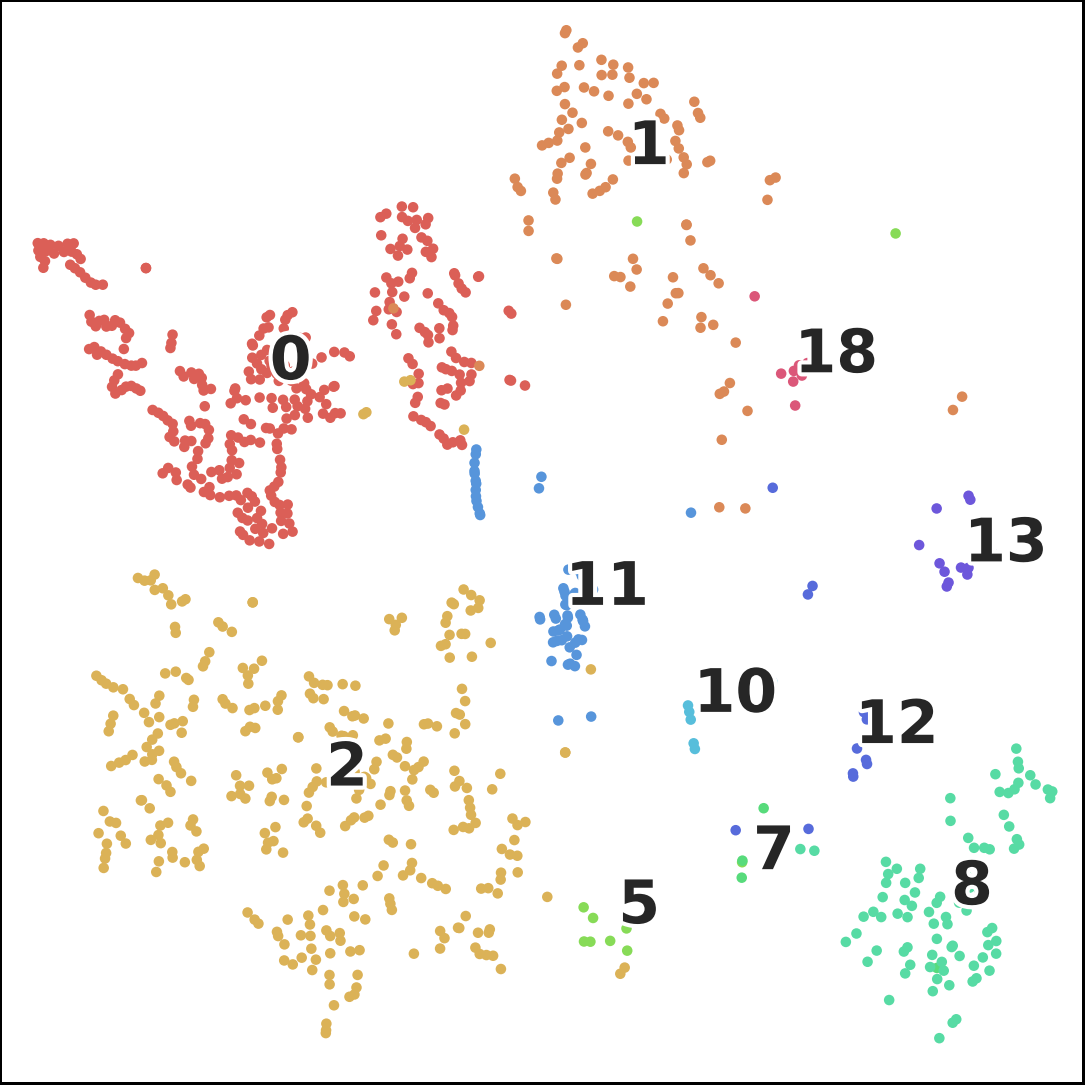}}
\vspace{-2.5 pt}
\\
\raisebox{-0.5\height}{S+T}
& \raisebox{-0.5\height}{IDA~\cite{pan2020unsupervised}}
& \raisebox{-0.5\height}{\textbf{SSDAS (Ours)}}
\\
\end{tabular}
\vspace{2.5 pt}
\caption{
\textbf{First row:} Qualitative illustration of semi-supervised domain adaptive semantic segmentation for GTA5 $\rightarrow$ Cityscapes adaptation. SSDAS employs a few labeled target sample as anchors for adaptive and progressive feature alignment between labeled source samples and unlabeled target samples, which produces nice semantic segmentation especially for the challenging low-frequency categories such as pole, bus and traffic-light etc. \textbf{Second row:} t-SNE \cite{maaten2008visualizing} visualization of feature distribution for target images in task GTA $\rightarrow$ Cityscapes: Each colour represents one semantic class of image pixels with a digit showing the class centre. $\sigma_{w}^{2}$ and $\sigma_{b}^{2}$ on the top of each graph are intra-class variance and inter-class distance of the corresponding feature distribution. The proposed SSDAS greatly outperforms ``S+T" and ``IDA" baselines in domain adaptive semantic segmentation qualitatively and quantitatively.
}
\label{fig:results}
\end{figure*}

\section{Experiments}
This section presents the evaluation of our SSDAS including datasets and implementation details, comparisons with the state-of-the-art, ablation studies, and discussion, more details to be described in the ensuing subsections.

\subsection{Experiment Setups}
In our experiments, we followed the setting of ~\cite{saito2019semi} that focuses on SSDA-based image classification. The training data consist of three parts including labeled source samples, unlabeled target samples, and 1 or 3 labeled target samples that are randomly selected for 1-shot or 3-shot SSDA-based semantic segmentation, respectively.

\textbf{Datasets:} We evaluated SSDAS over two challenging domain adaptive segmentation tasks GTA5$\rightarrow$Cityscapes and SYNTHIA$\rightarrow$Cityscapes, which involve two synthetic source datasets and one real target dataset. Specifically, Cityscapes consists of $2975$ training images and $500$ validation images. GTA5 and SYNTHIA consist of $24,966$ and $9,400$ high-resolution synthetic images which share 19 and 16 classes with Cityscapes. We also evaluated SSDAS on domain adaptive classification over the dataset Office-Home~\cite{venkateswara2017deep} that has 65 image classes of 4 domains.

\textbf{Implementation Details.} All our experiments were implemented in Pytorch. The segmentation model $G$ uses ResNet101 \cite{he2016deep} (pre-trained with ImageNet \cite{deng2009imagenet}) with DeepLab-V2~\cite{chen2017deeplab}. 
The optimizer is SGD \cite{bottou2010large} with a momentum of $0.9$ and a weight decay of $1e-4$. The learning rate is $2.5e-4$ initially and decreased by a polynomial policy with a power of $0.9$. Except parameter studies in Table~\ref{tab:abla_weight_jigsaw}, we set the trade-off parameter $\lambda_{j}$ at 0.1 and the number of Jigsaw Puzzle classes $N$ at 100 in all other experiments. The Jigsaw Puzzle classifiers (\ie, $\mathcal{J}_{s}$ and $\mathcal{J}_{t}$) are pre-trained with ``S+T" model to avoid noisy predictions at the initial training stage. We freeze Jigsaw Puzzle classifier $\mathcal{J}_{t}$ when its training loss is smaller than $\mathcal{J}_{s}$'s to avoid over-fitting with just a few labeled target samples. For SSDA-based classification, we follow the setting in~\cite{saito2019semi}.

\subsection{Ablation Studies}
We first examine different SSDAS components to study their contributions to SSDA-based semantic segmentation. Table \ref{tab:abla} shows experimental results over the validation set of Cityscapes, where the first row shows the result of ``S+T" model that is trained with supervised loss with labeled source and target samples only (with no alignment) as defined in Eq.~\ref{eq_S+T}. It can be seen that ``S+T" model does not perform well due to cross-domain gaps.

However, ACDA improves ``S+T" model clearly at both region and image levels, largely because ACDA employs few-shot target features to re-weight source features adaptively by increasing (or decreasing) the weight of target-alike (or target-unlike) source features. Further including PIDA improves mIoU by another $+1.4\%$ at both image and region levels, demonstrating its effectiveness in intra-domain alignment. It also shows that ACDA and PIDA are complementary by focusing on cross-domain alignment and intra-domain alignment, respectively. 

The last three rows show that including image-level alignment and region-level alignment simultaneously outperforms adopting either one alone for both ACDA and PIDA. This shows that our proposed image-level and region-level alignment are complementary, where the image-level alignment focuses more on classes with big sizes (\eg, road, building, sky, etc.) while the region-level alignment focuses more on classes with small sizes (\eg, person, car, bike, etc.). Finally, including both alignment strategies at both image and region levels (\ie, the complete SSDAS model) performs clearly the best.

\subsection{Comparisons with the State-of-Art}

As there is few prior work on SSDA-based semantic segmentation, we conducted two sets of experiments to benchmark our SSDAS with the state-of-the-art. In the first set of experiments, we adapted state-of-the-art UDA methods for the SSDA task. Specifically, we included the labeled target samples and the corresponding supervised loss into the UDA methods to approximate SSDA-based semantic segmentation. Tables \ref{table:gta2city} and \ref{table:synthia2city} show representative UDA methods (`AdaptSeg', `ADVENT', `CRST', `FDA' and `CrCDA') and their results. It can be seen that our SSDAS outperforms all adapted UDA methods consistently for both tasks GTA5$\rightarrow$Cityscapes and SYNTHIA$\rightarrow$Cityscapes. The superior segmentation is largely attributed to the adaptive and progressive feature alignment in SSDAS that exploits the few-shot labeled target samples to guide the cross-domain and intra-domain alignment effectively.

In the second set of experiments, we benchmarked SSDAS with state-of-the-art SSDA-based image classification methods for both semantic segmentation and image classification tasks. To adapt SSDAS for image classification, we simply take the feature maps instead of segmentation maps as the Jigsaw Puzzle input with little fine-tuning. Tables \ref{tab:SSDA_classification_comp1} and \ref{tab:SSDA_classification_comp2} show experimental results. For the semantic segmentation in Table \ref{tab:SSDA_classification_comp1}, we can see that SSDAS outperforms SSDA classification methods by large margins (over 5.3\% in mIoU) for both 1-shot and 3-shot settings. For the image classification task in Table \ref{tab:SSDA_classification_comp2}, SSDAS outperforms state-of-the-art SSDA classification methods clearly as well. These experiments show that SSDAS is generic for different tasks.

\subsection{Discussion}
\textbf{Number of labeled target samples:} We studied how SSDAS behaves while including more labeled target samples in training. The experiments in Table~\ref{tab:abla_number_labeled} shows that domain adaptive segmentation can be improved consistently when more labeled target samples are included. 

\renewcommand\arraystretch{1.1}
\begin{table}[t]
\centering
\scalebox{0.9}{
\begin{tabular}{cp{0.9cm}<{\centering}p{0.9cm}<{\centering}p{0.9cm}<{\centering}p{0.9cm}<{\centering}p{0.9cm}<{\centering}}
\hline
& \multicolumn{5}{c}{The number of the labeled target samples}
\\\hline
Method & \multicolumn{1}{c}{1} & \multicolumn{1}{c}{3} & \multicolumn{1}{c}{5} & \multicolumn{1}{c}{10} &\multicolumn{1}{c}{20}
\\\hline
S+T &37.9 &39.7 &40.4  &41.7  &44.6\\
\textbf{SSDAS} &48.5 &49.7 &50.1  &51.1  &52.6\\
\hline
\end{tabular}}
\caption{
The number of labeled target samples matters: Domain adaptive semantic segmentation keeps improving with the increase of labeled target samples (over the task GTA $\rightarrow$ Cityscapes).
}
\label{tab:abla_number_labeled}
\end{table}

\renewcommand\arraystretch{1.1}
\begin{table}[t]
\centering
\scalebox{0.9}{
\begin{tabular}{cccccc}
\hline
& \multicolumn{5}{c}{Parameter Analysis of $\lambda_{j}$ and $N$}
\\\toprule
$\lambda_{j}$ & \multicolumn{1}{c}{0.025} & \multicolumn{1}{c}{0.05} & \multicolumn{1}{c}{0.1} & \multicolumn{1}{c}{0.2} & \multicolumn{1}{c}{0.4}
\\\hline
SSDAS &47.9 &48.3 &48.5  &48.4  &48.1\\\bottomrule
$N$ & \multicolumn{1}{c}{30} & \multicolumn{1}{c}{50} & \multicolumn{1}{c}{100} & \multicolumn{1}{c}{300} & \multicolumn{1}{c}{500} 
\\\hline
SSDAS &48.2 &48.3 &48.5  &48.4  &48.4 
\\
\hline
\end{tabular}}
\caption{
The  weight parameter $\lambda_{j}$ and the number of Jigsaw Puzzle classes $N$ mater: Domain adaptive semantic segmentation is slightly affected by $\lambda_{j}$ but tolerant to $N$ (for GTA $\rightarrow$ Cityscapes).
}
\label{tab:abla_weight_jigsaw}
\end{table}

\begin{table}
\centering
\vspace{1mm}
\scalebox{0.9}{
\begin{tabular}{l|cc|c} \hline
 Method & Convention-feature  &Context-feature    &mIoU\\\hline
\multirow{3}{*}{SSDAS} &\multicolumn{1}{c}{\checkmark} & &47.1\\
& &\multicolumn{1}{c|}{\checkmark} &48.5	\\
&\multicolumn{1}{c}{\checkmark} &\multicolumn{1}{c|}{\checkmark} &49.7	\\
\hline
\end{tabular}}
\caption{
SSDAS is generic and can work for other features such as conventional features. For the task GTA $\rightarrow$ Cityscapes, we employ SSDAS to aligning the convention-feature and context-feature, and present the domain adaptive semantic segmentation performance in mIoU.}
\label{tab:SSDA_with_conven_F}
\end{table}

\textbf{The weight $\lambda_{j}$ and the number of Jigsaw Puzzle classes $N$:} Parameters $\lambda_{j}$ and $N$ control the weights of supervised and unsupervised losses and the difficulty of Jigsaw Puzzle, respectively. We studied the two parameters by changing $\lambda_{j}$ from $0$ to $1$ with a step of $1/6$ and setting $N$ at a few number as shown in Table~\ref{tab:abla_weight_jigsaw}. Experiments over the task GTA$\rightarrow$Cityscapes show that SSDAS is tolerant to both $\lambda_{j}$ and $N$. The major reason is that the adaptive and progressive learning in SSDAS can alleviate `negative alignment' with dissimilar features.

\textbf{Context vs conventional features:} We use context feature in this work but SSDAS can also work with conventional segmentation features, i.e. the features from the feature extractor $E$ without including the following Jigsaw Puzzle classifiers. To work with conventional features, the algorithm and optimization functions are the same (as with context features) except that the context features are replaced by conventional features. Please refer to Section \textcolor{red}{A.1.} of supplementary materials for details about the optimization functions and algorithms of aligning conventional features. Table \ref{tab:SSDA_with_conven_F} compares SSDAS while working with context and conventional features. We can observe that SSDAS can work with convention features well though the segmentation performance drops clearly. The performance drop is largely due to the fact that context features capture context information and are more effective in semantic segmentation. In addition, experiments show that context and conventional features are complementary while working together in domain adaptive semantic segmentation.

Moreover, we provide qualitative comparison and feature distribution visualization over the task GTA5$\rightarrow$Cityscapes in Fig.~\ref{fig:results}.
Due to the space limit, more discussion and qualitative comparisons over domain adaptive segmentation task are provided in the appendix.

\section{Conclusion}

In this work, we presented SSDAS, a Semi-Supervised Domain Adaptive image Segmentation network that employs a few labeled target samples as anchors for adaptive and progressive feature alignment between labeled source samples and unlabeled target samples. To the best of our knowledge, this is the first effort towards semi-supervised domain adaptive semantic segmentation (with few-shot target samples).
Extensive experiments demonstrate the superiority of our SSDAS over a number of baselines including UDA-based segmentation and SSDA-based classification methods. In addition, SSDAS is complementary and can be easily integrated with UDA-based methods with consistent improvements in segmentation. We will explore how to better make use of a few labeled target samples in SSDA-based semantic segmentation. In addition, we will also study how to extend the idea of our SSDAS to other computer vision tasks such as object detection and panoptic segmentation.

\section*{Acknowledgement}
This research was conducted in collaboration with Singapore Telecommunications Limited and supported/partially supported (delete as appropriate) by the Singapore Government through the Industry Alignment Fund - Industry Collaboration Projects Grant.

{\small
\bibliographystyle{ieee_fullname}
\bibliography{egbib}
}

\end{document}